\documentclass{article}

\usepackage{PRIMEarxiv}
\usepackage{cite}
\usepackage{amsmath,amssymb,amsfonts}
\usepackage{algorithmic}
\usepackage{textcomp}
\usepackage{xcolor}
\usepackage{subcaption}
\usepackage{balance}
\usepackage{todonotes}
\usepackage{multirow}%
\usepackage{booktabs}
\usepackage{comment}
\usepackage[utf8]{inputenc} % allow utf-8 input
\usepackage[T1]{fontenc}    % use 8-bit T1 fonts
\usepackage{hyperref}       % hyperlinks
\usepackage{url}            % simple URL typesetting
\usepackage{booktabs}       % professional-quality tables
\usepackage{amsfonts}       % blackboard math symbols
\usepackage{nicefrac}       % compact symbols for 1/2, etc.
\usepackage{microtype}      % microtypography
\usepackage{lipsum}
\usepackage{fancyhdr}       % header
\usepackage{graphicx}       % graphics
\graphicspath{{media/}}     % organize your images and other figures under media/ folder
\def\BibTeX{{\rm B\kern-.05em{\sc i\kern-.025em b}\kern-.08em
    T\kern-.1667em\lower.7ex\hbox{E}\kern-.125emX}}
\title{Learning to Assess the Reliability of Number-of-Runs Estimation in Stochastic Optimization
\thanks{The authors acknowledge the support of the Horizon Europe EU ERA Chair AutoLearn-SI (101187010), as well as the Slovenian Research Agency through program grant No. P2-0098, project grant No. J2-70078 and  No. GC-0001. } 
}

\author{
  Sara Gjorgjieva$^{1}$, 
  Eva Tuba$^{1}$, 
  and Tome Eftimov$^{1}$\\
  \\
  $^{1}$Jo\v{z}ef Stefan Institute, Ljubljana, Slovenia \\
  \\
  \texttt{sara.gjorgjieva@ijs.si, eva.tuba@ijs.si, tome.eftimov@ijs.si}
}

\begin{document}
\maketitle

\begin{abstract}
    In large-scale benchmarking of stochastic optimization algorithms, the key challenge is no longer whether repeated runs are needed for reliability, but how to determine when sufficient evidence has been collected without incurring unnecessary computational cost. We study a learning-based extension of a recent empirical online heuristic that adaptively estimates the required number of runs using outlier handling and skewness-based symmetry checks. Using annotated outcomes from 132{,}000 Nevergrad runs on COCO (24 problems in 20 dimensions, 10 instances each, 11 optimizers), we train classifiers on 23 statistical, energy-free, and shape and stability features to predict whether a run-number estimate is reliable, prioritizing detection of incorrect estimates via minority-class recall. We evaluate reliability prediction using a within-configuration learning setup, where models are trained and tested on data sharing the same optimizer. The results show that run-number reliability can be learned in a within-configuration scenario, enabling detection of unreliable estimates with high minority-class recall, although performance remains limited by the restricted data diversity within fixed configurations.
\end{abstract}


\section{Introduction}
Large-scale benchmarking of stochastic optimization algorithms faces a fundamental tension between experimental reliability and computational efficiency~\cite{bartz2020benchmarking}. Reliable performance assessment requires repeated executions, yet the cost of extensive repetition grows rapidly with the size of algorithm portfolios, benchmark suites, and problem dimensions. As benchmarking studies increasingly scale to millions of runs, the question is no longer whether repetitions are needed, but how to determine when enough evidence has been collected.

Current practice largely resolves this question through static design choices, fixing the number of runs per problem instance in advance (e.g., 15, 25, 30, or 50)~\cite{hansen2009real,liang2013problem, derrac2011practical}. While simple to implement, such uniform protocols implicitly assume homogeneous algorithm behavior across problems, an assumption that rarely holds in practice. Empirical evidence shows that variability can differ substantially across algorithm–problem combinations, making fixed repetition budgets either wasteful or insufficient depending on the case~\cite{vevcek2017influence}.

An empirical online method was recently proposed to estimate the required number of runs for single-objective stochastic optimizers on a per-instance basis~\cite{eftimov2025adaptive}. The method uses probabilistic heurstic to detect convergence and stop runs once stability is reached. Experiments across large algorithm portfolios and benchmark suites show that it correctly identifies the needed run count in most cases while reducing total executions by about 50\%, improving benchmarking efficiency and sustainability. However, in a subset of cases the estimated run count deviates from the ground truth by 5–25\%, and such inaccuracies can only be detected retrospectively after all runs are completed, limiting its applicability in fully online benchmarking scenarios.

Building on the empirical evidence in~\cite{eftimov2025adaptive}, this work addresses the limitation by introducing a learning-based extension. Previously identified inaccurate cases are used as annotated data to extract statistical, energy-free, shape, and stability features from run samples. These features train a supervised classifier that predicts whether a run-number estimation is reliable. This enables early detection of unreliable estimates, transforming the heuristic into a fully online, self-assessing system that supports real-time experimental decisions while preserving the efficiency and sustainability benefits of adaptive run estimation.

\noindent\textbf{Outline:} Section~\ref{sec:related_work} reviews related work, Section~\ref{sec:methodology} formulates the task as a binary classification problem, and Section~\ref{sec:experimental_design} describes the experimental settings, features, classifiers, and evaluation protocol. Results are presented in Section~\ref{sec:results}. Section~\ref{sec:conclusion} concludes the paper. All data and code required to reproduce the experiments are available at~\cite{zenodo18351075}.

\section{Related work}
\label{sec:related_work}

The number of runs in stochastic optimization experiments is critical for valid benchmarking. Traditional protocols use fixed repetition counts based on convention~\cite{hansen2009real,liang2013problem}. While larger samples increase statistical sensitivity~\cite{bartz2007experimental}, they also raise computational cost and may mask practical significance~\cite{vevcek2014chess,eftimov2019identifying}. Insufficient repetitions can distort results, especially in automated configuration and large-scale benchmarking~\cite{vermetten2022analyzing}. This has led to more principled methods linking run counts to statistical guarantees~\cite{campelo2019sample}, though these are mainly suited for offline experimental planning and often require explicit algorithm comparisons.

Another line of work explores adaptive experimentation, where runs are dynamically continued or stopped based on observed outcomes. An empirical online method estimates the required runs by monitoring the stability of performance samples from repeated executions on a fixed problem instance~\cite{eftimov2025adaptive}. After initial runs, results are centered around the sample mean and distribution symmetry is assessed as an indicator of representativeness. Robust outlier detection (e.g., IQR filtering, percentile trimming, modified z-score) is applied before evaluating symmetry. Skewness is then monitored, and additional runs are executed until it falls within a predefined threshold. The final estimation can later be labeled as accurate or inaccurate, enabling the creation of annotated data for supervised learning.

The present study differs from prior work by explicitly modeling the uncertainty of adaptive stopping decisions. By learning from previously observed estimation failures, we introduce a supervised mechanism that predicts the reliability of run-number estimations during execution. This positions the work at the intersection of adaptive experimental design and learning-based decision support, extending the scope of machine learning in optimization benchmarking beyond performance prediction toward experimental self-assessment.

\section{Methodology}
\label{sec:methodology}
We consider a labeled dataset consisting of instances where each instance represents a sequence of objective function values obtained from repeated independent executions of a single-objective continuous optimization algorithm on a fixed problem instance. Each sequence is associated with a binary label indicating whether the estimated number of algorithm runs is correct or incorrect. Because different optimization processes may require different numbers of runs, the sequences vary in length. To enable supervised learning with fixed-dimensional inputs, each sequence is transformed into a feature vector through a feature extraction procedure that summarizes statistical, energy-free, and shape- and stability-related characteristics of the observed objective values. Although the values originate from successive algorithm executions, the sequence is treated as an unordered sample reflecting stochastic performance variability rather than as a time series.

The learning task is to train a binary classifier that predicts whether a run-number estimation is correct based on these extracted features. Since the dataset may exhibit class imbalance, a resampling strategy is applied to the training data after the train–test split, while the test set remains unchanged to ensure unbiased evaluation. The classifier is trained by minimizing a standard classification loss over the resampled training data within a chosen family of models defined by their hyperparameters. Hyperparameter selection is guided by maximizing the recall of the minority class, which corresponds to cases where the estimation is incorrect. This choice prioritizes the detection of unreliable estimations, as false negatives—incorrect estimations predicted as correct—may prematurely stop additional algorithm runs and potentially affect the final optimization outcome. In contrast, false positives only lead to redundant runs without compromising correctness. The final model is therefore selected as the configuration that maximizes this minority-class recall criterion.

\section{Experimental design}
\label{sec:experimental_design}
\noindent\textbf{Data origin and annotation:} The experimental data used in this study originates from a previously published large-scale benchmarking study~\cite{eftimov2025adaptive} and is further annotated for this work. The original experiments estimated the required number of runs for reliable performance evaluation of stochastic optimizers. A post hoc analysis identified cases where these estimates were accurate or inaccurate, which serve as ground-truth labels in this study.

\noindent\textbf{Benchmark suite and problem setup: } All experiments use the \textit{COmparing Continuous Optimizers} (COCO) benchmark suite~\cite{hansen2021coco}, which includes 24 single-objective continuous optimization problem classes with multiple instances generated through shifting, scaling, and rotation. Although the original dataset considers dimensions $D \in \{10,20,40\}$, the Nevergrad-based experiments used here focus on $D = 20$, following the setup of the original study~\cite{nevergrad}.

\noindent\textbf{Algorithm portfolio: }The algorithm portfolio includes 11 optimization algorithms from the Nevergrad framework~\cite{nevergrad}: Differential Evolution (DE), Diagonal CMA, NaiveIsoEMNA, NGOpt14, NGOpt38, OnePlusOne, modCMA, modDE, Particle Swarm Optimization (PSO), Random Search, and RCobyla. All algorithms were run with default hyperparameters and a stopping criterion of $D \times 2{,}000$ function evaluations. For each algorithm--problem combination, 10 instances were evaluated with 50 independent runs per instance. In total, the experiments cover 240 problem–instance pairs, resulting in 132{,}000 algorithm runs, which serve as ground truth for evaluating adaptive run estimation.

\noindent\textbf{Outlier handling and skewness thresholds: } To reduce the influence of extreme stochastic outcomes, the original study considered three outlier detection methods: (i) interquartile range (IQR)~\cite{whaley2005interquartile}, (ii) percentile-based trimming~\cite{diciccio1988review}, and (iii) the modified z-score~\cite{kannan2015labeling}. Outliers are removed only for assessing symmetry of intermediate samples and do not affect the estimated number of runs. Symmetry is measured using sample skewness with thresholds of 0.05, 0.10, 0.15, and 0.20. The resulting combinations of outlier method and threshold were used to annotate the data reused in this study.

\noindent\textbf{Feature extraction: }
To capture both central tendencies and structural properties of the optimization behavior, a total of $23$ features are extracted from each objective-value sequence. These include \textit{first-order and dispersion statistics} (mean, median, standard deviation, variance, minimum, maximum, range, first and third quartiles ($q_{25}$, $q_{75}$), interquartile range, median absolute deviation, coefficient of variation), \textit{distributional measures} (skewness, kurtosis, entropy), and \textit{energy-related descriptors} (sum, sum of absolute values, signal energy, root mean square). In addition, \textit{shape and stability features} are computed to describe the optimization dynamics, including the slope of performance improvement, rolling standard deviation, fast Fourier transform (FFT) energy, and autocorrelation. These features are further associated with whether the early stopping decision is reliable.

\noindent\textbf{Model selection and hyperparameter optimization:} To explore supervised classifiers for reliability prediction, we performed hyperparameter optimization using the Optuna framework~\cite{akiba2019optuna}. The study considers several tree-based and ensemble models, including Decision Trees, Random Forests, XGBoost, LightGBM, and CatBoost, selected for their complementary bias–variance properties and robustness to heterogeneous feature distributions.

Hyperparameters were sampled from predefined ranges using Optuna. Given the strong class imbalance, the optimization objective maximizes $Recall_0$, prioritizing the detection of incorrect early stopping decisions. Hyperparameter tuning was performed only on training folds, with evaluation on held-out test sets. Initial broad search spaces led to overfitting and unstable generalization; therefore, the search was restricted to simpler and more robust configurations. All tested hyperparameter ranges and implementation details are available in our public code repository.

%E4 becomes E2, E2 E4, and E3 E5, E5 becomes E3
\noindent\textbf{Experimental setup: } To evaluate robustness and generalization of reliability prediction for run estimates, we consider a \textit{within-configuration learning} experiment. In this setting, models are trained and tested on data sharing the same outlier detection method, skewness threshold, and optimizer. Each combination defines a separate dataset (e.g., IQR, 0.05, modCMA), resulting in $3 \times 4 \times 11 = 132$ datasets. Each dataset contains 240 instances derived from 24 problems with 10 instances each. This setup evaluates reliability prediction within a fixed optimizer configuration, assuming prior annotations from the chosen threshold and outlier method.

\noindent\textbf{Implementation details:} We used a stratified 70/30 train–test split. Stratification is performed based on the binary outcome (0 for false estimation and 1 for true estimation), while also ensuring adequate representation of cases labeled as 1 when the maximum number of runs (50) specified in the annotation is reached. Feature values are standardized using z-score normalization, and the Synthetic Minority Over-sampling Technique (SMOTE) is applied on the training data to mitigate class imbalance. Next, we perform Optuna-based hyperparameter optimization using 5-fold cross-validation on SMOTE-enriched training data. The best model selected for each dataset is then evaluated on the held-out test set; these results are reported in the paper. The results from the 5-fold cross-validation and the corresponding model selection logs are provided in our repository. The logic here is to mimic a nested validation. 
\section{Results}
\label{sec:results}
\begin{figure*}
    \centering
    \includegraphics[width=0.9\linewidth]{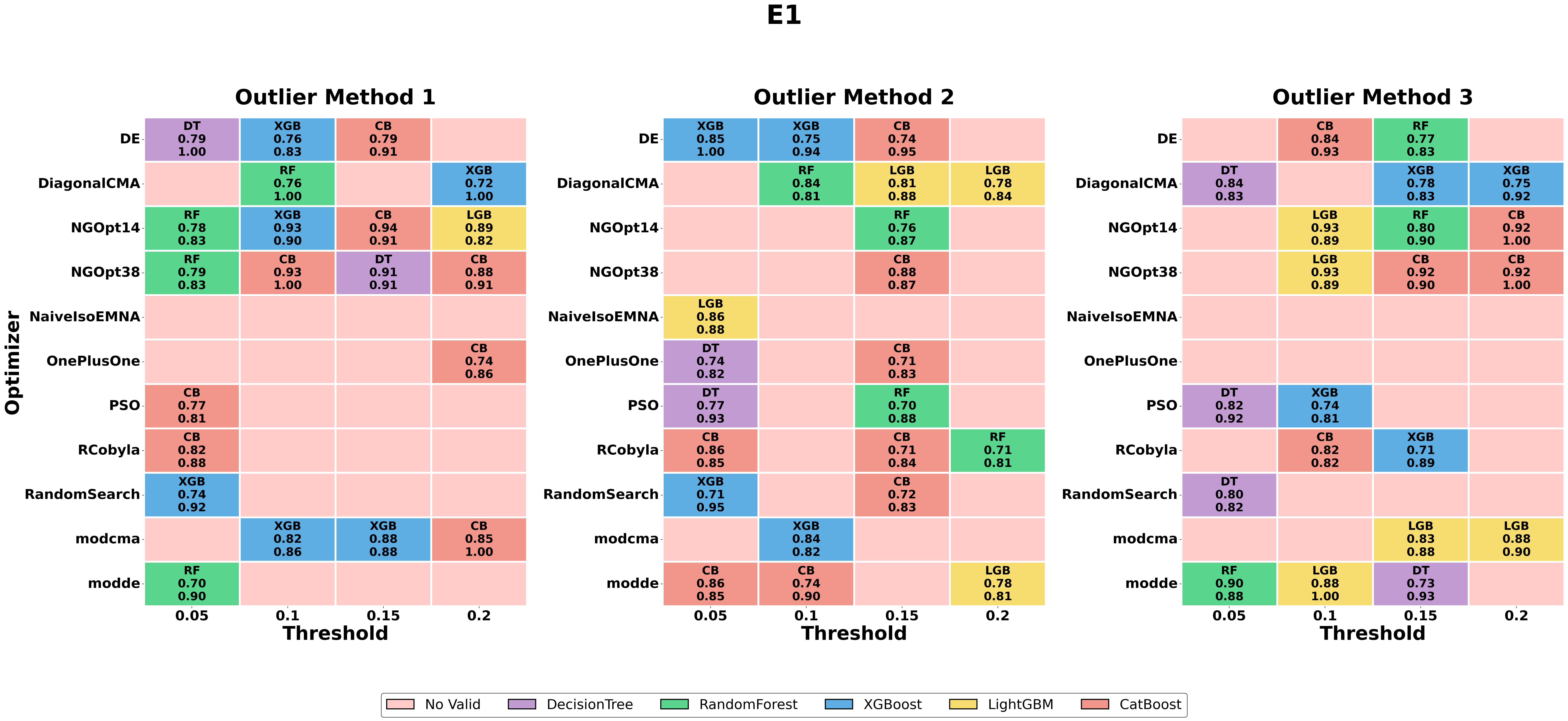}
    \caption{Within-configuration learning experiment results. Three heatmaps correspond to outlier methods~1,~2, and~3. Rows represent optimization algorithms, and columns represent skewness thresholds. Each colored cell reports the best classifier for configurations satisfying both criteria ($\mathrm{F1}_1 > 0.70$ and $\mathrm{Recall}_0 > 0.80$), showing the classifier abbreviation (first line), $\mathrm{F1}_1$ score (second line), and $\mathrm{Recall}_0$ score (third line). Cell colors indicate classifier type: purple (DecisionTree), green (RandomForest), blue (XGBoost), yellow (LightGBM), and salmon (CatBoost). Pink empty cells indicate that no valid classifier exists. Overall, only $48.5\%$ of configurations (64/132) yield valid models.
}
    \label{fig:E1}
\end{figure*}

To evaluate learnability of run-number reliability, we perform a within-configuration experiment. Due to strong class imbalance, we prioritize minority-class recall ($Recall_0$) to detect incorrect estimations and avoid premature stopping. Model selection maximizes $Recall_0$, while results analyze the trade-off with $F1_1$ to maintain performance on correct cases. For each configuration, we check whether at least one classifier meets $F1_1 > 0.70$ and $Recall_0 > 0.80$, prioritizing detection of unreliable estimates while maintaining acceptable accuracy. As shown in Figure~\ref{fig:E1}, valid classifiers are obtained in only $48.5\%$ of configurations (64/132), highlighting the difficulty of learning reliable models within a fixed setting.

An analysis by optimization algorithm reveals pronounced variability in predictability. DE, DiagonalCMA, NGOpt14, and NGOpt38 achieve the highest success rates, each yielding valid models in eight out of 12 configurations (66.7\%). In contrast, NaiveIsoEMNA is the most difficult to predict, with only one valid configuration out of 12 (8.3\%). The remaining algorithms exhibit intermediate performance: modde (58.3\%), modcma and RCobyla (50.0\%), PSO (41.7\%), RandomSearch (33.3\%), and OnePlusOne (25.0\%).

When comparing classifier performance against the majority-class baseline (see Table~\ref{tab:e1-results}), the results reveal an important trade-off. Only nine out of 132 datasets ( i.e., a combination of outlier method and a threshold) (6.8\%) produce a classifier that outperforms the baseline in terms of $F1_1$. These cases are concentrated almost exclusively on NGOpt14 and NGOpt38, each exceeding the baseline in four datasets (33.3\%), with RCobyla doing so in only a single dataset. However, it is critical to note that the baseline achieves $Recall_0 = 0.00$ in all cases, meaning that it completely fails to identify unreliable estimates, which constitutes the primary objective of the proposed models. While the baseline attains a high $F1_1$ by trivially predicting all instances as belonging to the majority (reliable) class, this strategy provides no practical utility for detecting problematic convergence cases. In contrast, the trained classifiers, despite frequently underperforming the baseline in terms of $F1_1$, consistently achieve substantially higher $Recall_0$ values (typically above 0.80), successfully identifying unreliable estimates that the baseline entirely misses.

These results reveal a key limitation of within-configuration learning: with only 240 training instances, models lack sufficient diversity to learn beyond majority-class prediction.

\begin{table}[!th]
\centering
\caption{Results outperforming the baseline model.}
\label{tab:e1-results}
\resizebox{.45\textwidth}{!}{% <------ Don't forget this %
\begin{tabular}{@{}llllllll@{}}
\toprule
\textbf{Dataset} & \textbf{Model} & \textbf{Pr$_1$} & \textbf{Rc$_1$} & \textbf{F1$_1$} & \textbf{Pr$_0$} & \textbf{Rc$_0$} & \textbf{F1$_0$} \\
\midrule
1\_0.15\_NGOpt14 & CatBoost  & 0.982 & 0.902 & 0.940 & 0.625 & 0.909 & 0.741 \\
1\_0.1\_NGOpt14  & XGBoost   & 0.982 & 0.887 & 0.932 & 0.563 & 0.900 & 0.692 \\
1\_0.1\_NGOpt14  & CatBoost  & 1.000 & 0.871 & 0.931 & 0.556 & 1.000 & 0.714 \\
1\_0.1\_NGOpt38  & CatBoost  & 1.000 & 0.871 & 0.931 & 0.556 & 1.000 & 0.714 \\
2\_0.1\_NGOpt38  & XGBoost   & 0.930 & 0.914 & 0.922 & 0.667 & 0.714 & 0.690 \\
2\_0.1\_NGOpt38  & LightGBM  & 0.944 & 0.879 & 0.911 & 0.611 & 0.786 & 0.688 \\
3\_0.05\_RCobyla & LightGBM  & 0.931 & 0.900 & 0.915 & 0.571 & 0.667 & 0.615 \\
3\_0.1\_NGOpt14  & XGBoost   & 0.966 & 0.905 & 0.934 & 0.538 & 0.778 & 0.636 \\
3\_0.1\_NGOpt14  & CatBoost  & 0.967 & 0.937 & 0.952 & 0.636 & 0.778 & 0.700 \\
3\_0.1\_NGOpt38  & CatBoost  & 0.967 & 0.921 & 0.943 & 0.583 & 0.778 & 0.667 \\
3\_0.2\_NGOpt14  & CatBoost  & 1.000 & 0.852 & 0.920 & 0.550 & 1.000 & 0.710 \\
3\_0.2\_NGOpt38  & CatBoost  & 1.000 & 0.852 & 0.920 & 0.550 & 1.000 & 0.710 \\
\bottomrule
\end{tabular}% <------ Don't forget this %
}
\end{table}

\section{Conclusion}
\label{sec:conclusion}
This study evaluates the learnability of run-number reliability using a within-configuration experiment. The results show that classifiers can detect unreliable estimates with high minority-class recall, although valid models are obtained in only $48.5\%$ of configurations. While the majority-class baseline achieves higher $F1_1$ in most cases, it fails to detect unreliable estimates, whereas the learned classifiers successfully identify them. These findings demonstrate that reliability prediction is possible but challenging in data-limited settings, where restricted sample diversity limits model performance. Further work will investigate alternative experimental scenarios, including settings where data from different optimization algorithms are combined to increase training diversity.

\section*{Acknowledgment}
The authors acknowledge the support of the Horizon Europe EU ERA Chair AutoLearn-SI (101187010), as well as the Slovenian Research Agency through program grant No. P2-0098, project grant No. J2-70078 and  No. GC-0001.

\end{document}